\documentclass[11pt, a4paper, logo]{deepmind}

\usepackage{hyperref}
\usepackage{nicefrac}       
\usepackage{amsthm}
\usepackage{subcaption}
\usepackage{multirow}
\usepackage{multicol}
\usepackage{microtype}  
\usepackage[title]{appendix}

\usepackage{subfiles}
\usepackage{enumerate}
\usepackage{enumitem}
\usepackage{cancel}
\usepackage{tikz}
\usetikzlibrary{positioning,shapes,backgrounds}

\usepackage{natbib}
\theoremstyle{definition}

\newtheorem{proposition}{Proposition}
\newtheorem{example}{Example}

\newcommand{\rollout}[3]{{#1}^{(#2)}_{#3}}
\newcommand{\inaction}{\rollout{s}{0}{t}}
\newcommand{\base}{s'_t}
\newcommand{\cur}{s_t}
\newcommand{\start}{s_0}
\newcommand{\prev}{s_{t-1}}
\newcommand{\nxt}{s_{t+1}}
\newcommand{\step}{\rollout{s}{t-1}{t}}

\newcommand{\trans}{p}
\newcommand{\rew}{r}
\newcommand{\pair}[2]{(#1;#2)}
\newcommand{\act}{a_t}
\newcommand{\states}{\mathcal{S}}
\newcommand{\actions}{\mathcal{A}}
\newcommand{\rewards}{\mathcal{R}}
\newcommand{\noop}{\text{noop}}
\newcommand{\devc}[2]{d\pair{#1}{#2}}
\newcommand{\devx}[3]{d_{#1}\pair{#2}{#3}}
\newcommand{\dev}[1]{d_{#1}\pair{\cur}{\base}}
\newcommand{\reach}[2]{R\pair{#1}{#2}}
\newcommand{\ns}[3]{N_{#1}\pair{#2}{#3}}
\newcommand{\gammar}{\gamma_r}
\newcommand{\tr}[1]{\max(#1, 0)}
\newcommand{\rv}[2]{RV\pair{#1}{#2}}

\newcommand{\sizeA}{0.1}
\newcommand{\sizeB}{0.15}
\definecolor{lightblue}{HTML}{00aeff}
\definecolor{darkblue}{HTML}{191970}
\definecolor{seagreen}{HTML}{3CB371}
\newcommand{\xmark}{\textcolor{red}{\sffamily X}}

\DeclareMathOperator*{\expect}{\mathbb{E}}
\DeclareMathOperator*{\reals}{\mathbb{R}}

\title{Penalizing side effects using stepwise relative reachability}

\correspondingauthor{vkrakovna@google.com}

\author[1]{Victoria Krakovna}
\author[1]{Laurent Orseau}
\author[1]{Ramana Kumar}
\author[1]{Miljan Martic}
\author[1]{Shane Legg}

\affil[1]{DeepMind}

\paperurl{https://arxiv.org/abs/1806.01186}

\begin{abstract}
How can we design safe reinforcement learning agents that avoid unnecessary disruptions to their environment? We show that current approaches to penalizing side effects can introduce bad incentives, e.g. to prevent any irreversible changes in the environment, including the actions of other agents. To isolate the source of such undesirable incentives, we break down side effects penalties into two components: a baseline state and a measure of deviation from this baseline state. We argue that some of these incentives arise from the choice of baseline, and others arise from the choice of deviation measure. We introduce a new variant of the stepwise inaction baseline and a new deviation measure based on relative reachability of states. The combination of these design choices avoids the given undesirable incentives, while simpler baselines and the unreachability measure fail. We demonstrate this empirically by comparing different combinations of baseline and deviation measure choices on a set of gridworld experiments designed to illustrate possible bad incentives.
\end{abstract}

\begin{document}
\maketitle
\balance

\newif\ifarxiv
\arxivtrue

\newif\ifextensions
\extensionstrue
\extensionsfalse

\ifarxiv
  \section{Introduction}\label{sec:intro}
\else
  \section{INTRODUCTION}\label{sec:intro}
\fi

An important component of safe behavior for reinforcement learning agents is avoiding unnecessary side effects while performing a task~\citep{AmodeiOlah16,Taylor16}. For example, if an agent's task is to carry a box across the room, we want it to do so without breaking vases, while an agent tasked with eliminating a computer virus should avoid unnecessarily deleting files.
The side effects problem is related to the frame problem in classical AI~\citep{Mccarthy69}. For machine learning systems, it has mostly been studied in the context of safe exploration during the agent's learning process~\citep{Pecka14,Garcia15}, but can also occur after training if the reward function is misspecified and fails to penalize disruptions to the environment~\citep{Ortega18}. 

We would like to incentivize the agent to avoid side effects without explicitly penalizing every possible disruption, defining disruptions in terms of predefined state features, or going through a process of trial and error when designing the reward function. While such approaches can be sufficient for agents deployed in a narrow set of environments, they often require a lot of human input and are unlikely to scale well to increasingly complex and diverse environments. It is thus important to develop more general and systematic approaches for avoiding side effects. 


Most of the general approaches to this problem are reachability-based methods:
safe exploration methods that preserve reachability of a starting state~\citep{Moldovan12,Eysenbach17}, and reachability analysis methods that require reachability of a safe region~\citep{Mitchell05,Gillula12,Fisac17}. The reachability criterion has a notable limitation: it is insensitive to the magnitude of the irreversible disruption, e.g. it equally penalizes the agent for breaking one vase or a hundred vases, which results in bad incentives for the agent.
Comparison to a starting state also introduces undesirable incentives in \emph{dynamic} environments, where irreversible transitions can happen spontaneously (due to the forces of nature, the actions of other agents, etc). Since such transitions make the starting state unreachable, the agent has an incentive to interfere to prevent them. This is often undesirable, e.g. if the transition involves a human eating food.
Thus, while these methods address the side effects problem in environments where the agent is the only source of change and the objective does not require irreversible actions, a more general criterion is needed when these assumptions do not hold.

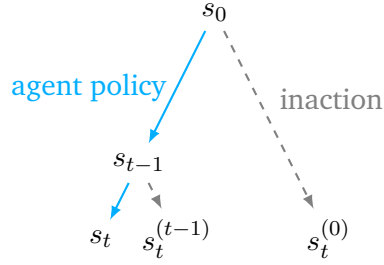
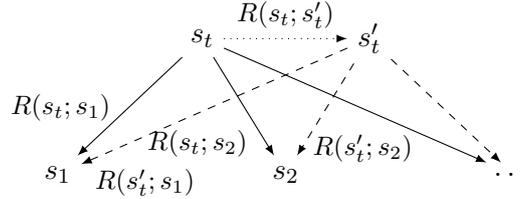
\begin{figure}[t]
\begin{subfigure}{\columnwidth}
\centering
    \begin{tikzpicture}[auto]
    \node (s0) at (0, 3) {$\start$};
    \node (st-1) at (-1, 1) {$\prev$};
    \node (st) at (-1.5, 0) {$\cur$};
    \node (step) at (-0.5, 0) {$\step$};
    \node (in) at (1.5, 0) {$\inaction$};
    \draw (s0) edge[->, >=latex, thick, color=lightblue] node[left]{agent policy} (st-1);
    \draw (st-1) edge[->, >=latex, thick, color=lightblue]  (st);
    \draw (s0) edge[->, >=latex, thick, dashed, color=gray] node{inaction}  (in);
    \draw (st-1) edge[->, >=latex, thick, dashed, color=gray]  (step);
    \end{tikzpicture}
  \caption{Choices of baseline state $\base$: starting state $\start$, inaction $\inaction$, and stepwise inaction $\step$. Actions drawn from the agent policy are shown by solid blue arrows, while actions drawn from the inaction policy are shown by dashed gray arrows.
  \label{fig:baselines}}
  \end{subfigure}
\begin{subfigure}{\columnwidth}
\centering
    \begin{tikzpicture}[auto]
        \node (sc) at (-1.1, 1.8) {$\cur$};
        \node (sb) at (1.1, 1.8) {$\base$};
        \node (s1) at (-3, 0) {$s_1$};
        \node (s2) at (0, 0) {$s_2$};
        \node (s3) at (3, 0) {$\dots$};
        \draw (sc) edge[->,>=latex] node[shift={(-1.7,0.3)}]{\small $\reach{\cur}{s_1}$} (s1);
        \draw (sc) edge[->,>=latex] node[shift={(-1.4,-.8)}]{\small $\reach{\cur}{s_2}$} (s2);
        \draw (sc) edge[->,>=latex] (s3);
        \draw (sb) edge[->,>=latex, dashed] node[shift={(-1.7,-.7)}]{\small $\reach{\base}{s_1}$} (s1);
        \draw (sb) edge[->,>=latex, dashed] node[shift={(-0.3,-0.2)}]{\small $\reach{\base}{s_2}$} (s2);
        \draw (sb) edge[->,>=latex, dashed] (s3);
        \draw (sc) edge[->,>=latex, dotted] node{\small $\reach{\cur}{\base}$} (sb);
    \end{tikzpicture}
    \caption{Choices of deviation measure $d$: given a state reachability function $R$, $\dev{UR} := 1-\reach{\cur}{\base}$ is the \emph{unreachability} measure of the baseline state $\base$ from the current state $\cur$ (dotted line), while relative reachability $\dev{RR} := \frac{1}{|\states|} \sum_{s\in \states} \tr{\reach{\base}{s} - \reach{\cur}{s}}$ is defined as the average reduction in reachability of states $s=s_1, s_2, \dots$ from current state $\cur$ (solid lines) compared to the baseline state $\base$ (dashed lines). \label{fig:rr}}
\end{subfigure}
\caption{Design choices for a side effects penalty: baseline states and deviation measures.}
\label{fig:breakdown}
\end{figure}

The contributions of this paper are as follows. In Section \ref{sec:components}, we introduce a breakdown of side effects penalties into two design choices, a baseline state and a measure of deviation of the current state from the baseline state, as shown in Figure \ref{fig:breakdown}. We outline several possible bad incentives (interference, offsetting, and magnitude insensitivity) and introduce toy environments that test for them. We argue that interference and offsetting arise from the choice of baseline, while magnitude insensitivity arises from the choice of deviation measure. 
In Section \ref{sec:baselines}, we propose a variant of the \emph{stepwise inaction} baseline, shown in Figure \ref{fig:baselines}, which avoids interference and offsetting incentives. In Section \ref{sec:deviation}, we propose a \emph{relative reachability} measure that is sensitive to the magnitude of the agent's effects, which is defined by comparing the reachability of states between the current state and the baseline state, as shown in Figure \ref{fig:rr}. 
\ifarxiv
(The relative reachability measure was originally introduced in the first version of this paper.) We also compare to the \emph{attainable utility} measure~\citep{Turner19}, which generalizes the relative reachability measure. 
\fi
In Section \ref{sec:experiments}, we compare all combinations of the baseline and deviation measure choices from Section \ref{sec:components}. We show that the unreachability measure produces the magnitude insensitivity incentive for all choices of baseline, 
\ifarxiv
while the relative reachability and attainable utility measures with the stepwise inaction baseline avoid the three undesirable incentives. 
\else
while the relative reachability measure with the stepwise inaction baseline avoids the three undesirable incentives. 
\fi

We do not claim this approach to be a complete solution to the side effects problem, since there may be other cases of bad incentives that we have not considered. However, we believe that avoiding the bad behaviors we described is a bare minimum for an agent to be both safe and useful, so our approach provides some necessary ingredients for a solution to the problem. 


\ifarxiv
\subsection{Preliminaries}\label{sec:setup}
\else
\subsection{PRELIMINARIES}\label{sec:setup}
\fi

We assume that the environment is a discounted Markov Decision Process (MDP), defined by a tuple $(\states, \actions, \rew, \trans, \gamma)$. $\states$ is the set of states, $\actions$ is the set of actions, $r: \states \times \actions \rightarrow \reals$ is the reward function, $\trans(\nxt|\cur, a_t)$ is the transition function, and  $\gamma \in (0,1)$ is the discount factor.

At time step $t$, the agent receives the state $\cur$, outputs the action $\act$ drawn from its policy $\pi(\act|\cur)$, and receives reward $\rew(\cur, \act)$. We define a transition as a tuple $(\cur, \act, \nxt)$ consisting of state $\cur$, action $\act$, and next state $\nxt$. 
We assume that there is a special \emph{noop} action $a^\noop$ that has the same effect as the agent being turned off during the given time step.

\ifarxiv
\subsection{Intended effects and side effects}
\else
\subsection{INTENDED EFFECTS AND SIDE EFFECTS}
\fi

We begin with some motivating examples for distinguishing intended and unintended disruptions to the environment:


\begin{example}[Vase]
\label{ex:vase}
The agent's objective is to get from point A to point B as quickly as possible, and there is a vase in the shortest path that would break if the agent walks into it.
\end{example}

\begin{example}[Omelette]
\label{ex:omelette}
The agent's objective is to make an omelette, which requires breaking some eggs.
\end{example}

In both of these cases, the agent would take an irreversible action by default (breaking a vase vs breaking eggs). However, the agent can still get to point B without breaking the vase (at the cost of a bit of extra time), but it cannot make an omelette without breaking eggs. We would like to incentivize the agent to avoid breaking the vase while allowing it to break the eggs.

Safety criteria are often implemented as constraints~\citep{Garcia15,Moldovan12,Eysenbach17}. This approach works well if we know exactly what the agent must avoid, but is too inflexible for a general criterion for avoiding side effects. For example, a constraint that the agent must never make the starting state unreachable would prevent it from making the omelette in Example \ref{ex:omelette}, no matter how high the reward for doing so.

A more flexible way to implement a side effects criterion is by adding a penalty for impacting the environment to the reward function, which acts as an intrinsic pseudo-reward. An impact penalty at time $t$ can be defined as a measure of \emph{deviation} of the current state $\cur$ from a \emph{baseline} state $\base$, denoted as $\devc{\cur}{\base}$. 
Then at every time step $t$, the agent receives the following total reward:
$$\rew(\cur,\act) - \beta \cdot \devc{\nxt}{\nxt'}.$$
Since the task reward $r$ indicates whether the agent has achieved the objective, we can distinguish intended and unintended effects by balancing the task reward and the penalty using the scaling parameter $\beta$. Here, the penalty would outweigh the small reward gain from walking into the vase over going around the vase, but it would not outweigh the large reward gain from breaking the eggs. 

\ifarxiv
\section{Design choices for an impact penalty}\label{sec:components}
\else
\section{DESIGN CHOICES FOR AN IMPACT PENALTY}\label{sec:components}
\fi


When defining the impact penalty, the baseline $\base$ and deviation measure $d$ can be chosen separately.
We will discuss several possible choices for each of these components.

\ifarxiv
\subsection{Baseline states}\label{sec:baselines}
\else
\subsection{BASELINE STATES}\label{sec:baselines}
\fi

\textbf{Starting state baseline.} 
One natural choice of baseline state is the starting state $\base=\start$ when the agent was deployed (or a starting state distribution), which we call the \emph{starting state baseline}. This is the baseline used in reversibility-preserving safe exploration approaches, where the agent learns a reset policy that is rewarded for reaching states that are likely under the initial state distribution.

\begin{figure}[ht]
\centering
\includegraphics[scale=.4]{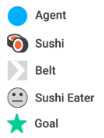}
\includegraphics[scale=.23]{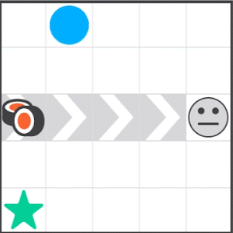}
\caption{Sushi environment.}
\label{fig:sushi_gridworld}
\end{figure}

While penalties with the starting state baseline work well in environments where the agent is the only source of change, in dynamic environments they also penalize irreversible transitions that are not caused by the agent. This incentivizes the agent to interfere with other agents and environment processes to prevent these irreversible transitions. To illustrate this \emph{interference} behavior, we introduce the Sushi environment, shown in Figure \ref{fig:sushi_gridworld}.

This environment is a Conveyor Belt Sushi restaurant. It contains a conveyor belt that moves to the right by one square after every agent action. There is a sushi dish on the conveyor belt that is eaten by a hungry human if it reaches the end of the belt. The interference behavior is to move the sushi dish off the belt (by stepping into the square containing the sushi). The agent is rewarded for reaching the goal square, and it can reach the goal with or without interfering with the sushi in the same number of steps. The desired behavior is to reach the goal without interference, by going left and then down.
An agent with no penalty performs well in this environment, but as shown in Section \ref{sec:experiments}, impact penalties with the starting state baseline produce the interference behavior.

\textbf{Inaction baseline.} 
Another choice is the \emph{inaction} baseline $\base=\inaction$: a counterfactual state of the environment if the agent had done nothing for the duration of the episode. Inaction can be defined in several ways. \cite{Armstrong17} define it as the agent never being deployed: conditioning on the event $X$ where the AI system is never turned on. It can also be defined as following some baseline policy, e.g. a policy that always takes the noop action $a^\noop$. We use this noop policy as the inaction baseline. 

\begin{figure}[ht]
\centering
\includegraphics[scale=\sizeB]{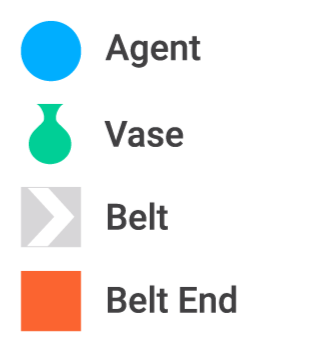}
\includegraphics[scale=\sizeB]{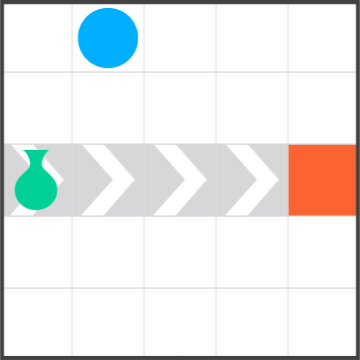}
\caption{Vase environment.}
\label{fig:vase_gridworld}
\end{figure}

Penalties with this baseline do not produce the interference behavior in dynamic environments, since transitions that are not caused by the agent would also occur in the counterfactual where the agent does nothing, and thus are not penalized. However, the inaction baseline incentivizes another type of undesirable behavior, called \emph{offsetting}. We introduce a Vase environment to illustrate this behavior, shown in Figure \ref{fig:vase_gridworld}.

\begin{figure}[ht]
\subcaptionbox{Agent takes the vase off the belt.}[\columnwidth]{
     \centering
     \includegraphics[scale=\sizeA]{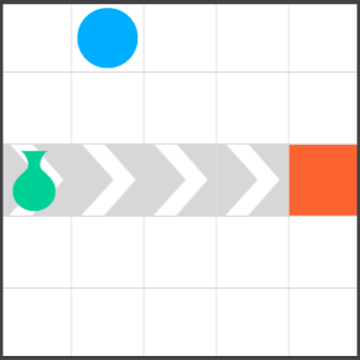}
     \includegraphics[scale=\sizeA]{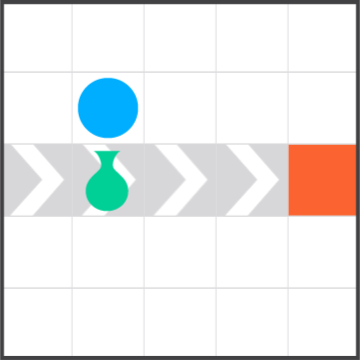}
     \includegraphics[scale=\sizeA]{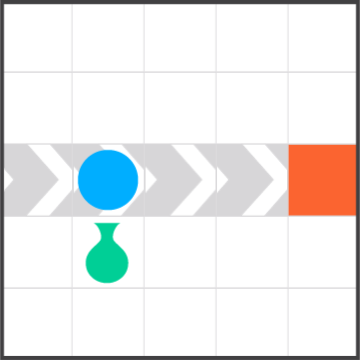}}
\subcaptionbox{Agent goes around the vase.}[\columnwidth]{
     \centering
     \includegraphics[scale=\sizeA]{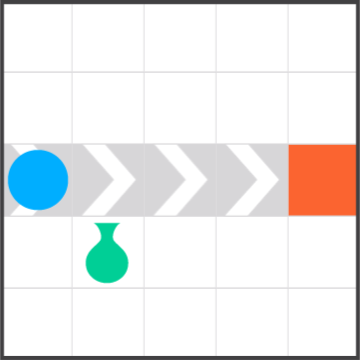}
     \includegraphics[scale=\sizeA]{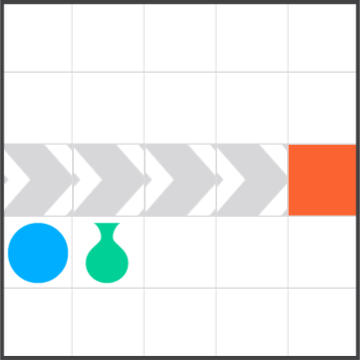}
     \includegraphics[scale=\sizeA]{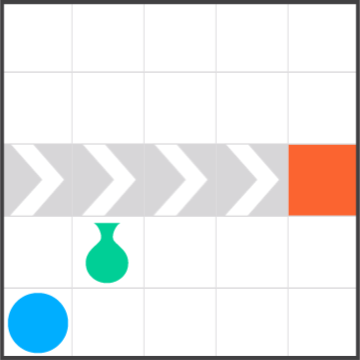}
     \includegraphics[scale=\sizeA]{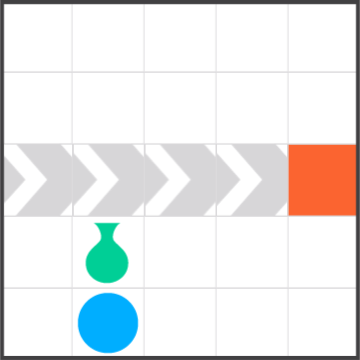}}
\subcaptionbox{Agent puts the vase back on the belt.}[\columnwidth]{
     \centering
     \includegraphics[scale=\sizeA]{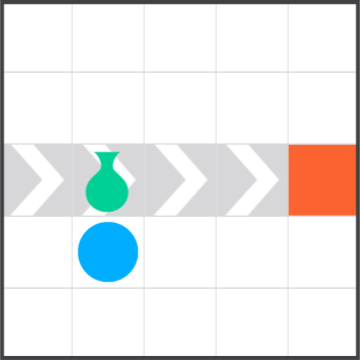}
     \includegraphics[scale=\sizeA]{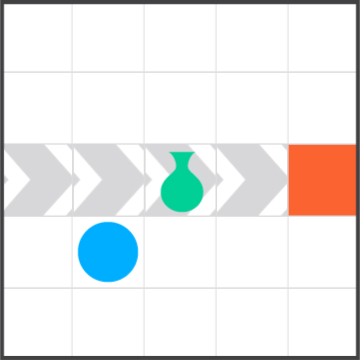}
     \includegraphics[scale=\sizeA]{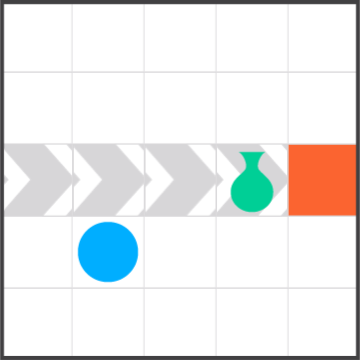}
     \includegraphics[scale=\sizeA]{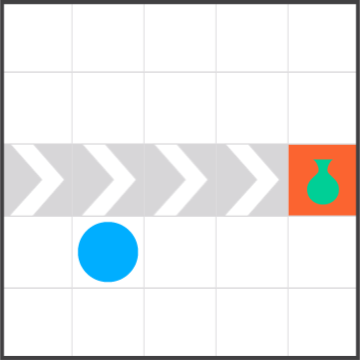}}
\caption{Offsetting behavior in the Vase environment.}
\label{fig:offsetting}
\end{figure}

This environment also contains a conveyor belt, with a vase that will break if it reaches the end of the belt. The agent receives a reward for taking the vase off the belt. The desired behavior is to move the vase off and then stay put. The offsetting behavior is to move the vase off (thus collecting the reward) and then put it back on, as shown in Figure \ref{fig:offsetting}.

Offsetting happens because the vase breaks in the inaction counterfactual. Once the agent takes the vase off the belt, it continues to receive penalties for the deviation between the current state and the baseline. Thus, it has an incentive to return to the baseline by breaking the vase after collecting the reward. Experiments in Section \ref{sec:experiments} show that impact penalties with the inaction baseline produce the offsetting behavior if they have a nonzero penalty for taking the vase off the belt.

\textbf{Stepwise inaction baseline.}
The inaction baseline can be modified to branch off from the previous state $\prev$ rather than the starting state $\start$. This is the \emph{stepwise inaction} baseline $\base=\step$: a counterfactual state of the environment if the agent had done nothing instead of its last action~\citep{Turner19}. 
This baseline state is generated by a baseline policy that follows the agent policy for the first $t-1$ steps, and takes an action drawn from the inaction policy (e.g. the noop action $a^\noop$) on step $t$. Each transition is penalized only once, at the same time as it is rewarded, so there is no offsetting incentive.

\begin{figure}[ht]
\centering
\begin{tikzpicture}[auto]
    \node (cur) at (-1.5, 0) {$\cur$};
    \node (base) at (-0.5, 0) {$\base$};
    \node (base1) at (0, -1) {$\rollout{s'}{t}{t+1}$};
    \node (base2) at (0.5, -2) {$\rollout{s'}{t}{t+2}$};
    \node (basedots) at (1, -3) {$\dots$};
    \node (basevase) at (-0.1, 0) {\includegraphics[scale=\sizeA]{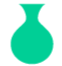}};
    \node (base1vase) at (0.5, -1) {\includegraphics[scale=\sizeA]{environments/vase.png}};
    \node (base2vase) at (1, -2) {\includegraphics[scale=\sizeA]{environments/vase.png}};
    \node (cur1) at (-1, -1) {$\rollout{s}{t}{t+1}$};
    \node (cur2) at (-0.5, -2) {$\rollout{s}{t}{t+2}$};
    \node (curdots) at (0, -3) {$\dots$};
    \node (curvase) at (-1.9, 0) {\includegraphics[scale=\sizeA]{environments/vase.png}};
    \node (cur1vase) at (-1.5, -1) {\includegraphics[scale=\sizeA]{environments/vase.png}};
    \node (cur2vase) at (-1, -2) {\includegraphics[scale=\sizeA]{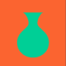}};
    \draw (base) edge[->, >=latex, thick, dashed, color=gray]  (base1);
    \draw (base1) edge[->, >=latex, thick, dashed, color=gray]  (base2);
    \draw (base2) edge[->, >=latex, thick, dashed, color=gray]  (basedots);
    \draw (cur) edge[->, >=latex, thick, dashed, color=gray]  (cur1);
    \draw (cur1) edge[->, >=latex, thick, dashed, color=gray]  (cur2);
    \draw (cur2) edge[->, >=latex, thick, dashed, color=gray]  (curdots);
\end{tikzpicture}
\caption{Inaction rollouts from the current state $\cur$ and baseline state $\base$ used for penalizing delayed effects of the agent's actions. If action $a_{t-1}$ puts a vase on a conveyor belt, then the vase breaks in the inaction rollout from $\cur$ but not in the inaction rollout from $\base$.} \label{fig:rollouts}
\end{figure}

However, there is a problem with directly comparing current state $\cur$ with $\step$: this does not capture delayed effects of action $a_{t-1}$. For example, if this action is putting a vase on a conveyor belt, then the current state $\cur$ contains the intact vase, and by the time the vase breaks, the broken vase will be part of the baseline state. Thus, the penalty for action $a_{t-1}$ needs to be modified to take into account future effects of this action, e.g. by using \emph{inaction rollouts} from the current state and the baseline (Figure \ref{fig:rollouts}).

An inaction rollout from state $\tilde \cur \in \{\cur, \base\}$ is a sequence of states obtained by following the inaction policy starting from that state: $\tilde \cur, \rollout{\tilde s}{t}{t+1}, \rollout{\tilde s}{t}{t+2}, \dots$. Future effects of action $a_{t-1}$ can be modeled by comparing an inaction rollout from $\cur$ to an inaction rollout from $\step$. For example, if action $a_{t-1}$ puts the vase on the belt, and the vase breaks 2 steps later, then $\rollout{s}{t}{t+2}$ will contain a broken vase, while $\rollout{s'}{t}{t+2}$ will not.
\cite{Turner19} compare the inaction rollouts $\rollout{s}{t}{t+k}$ and $\rollout{s'}{t}{t+k}$ at a single time step $t+k$, which is simple to compute, but does not account for delayed effects that occur after that time step. We will introduce a recursive formula for comparing the inaction rollouts $\rollout{s}{t}{t+k}$ and $\rollout{s'}{t}{t+k}$ for all $k\geq 0$ in Section \ref{sec:deviation}.

\ifarxiv
\subsection{Deviation measures}\label{sec:deviation}
\else
\subsection{DEVIATION MEASURES}\label{sec:deviation}
\fi

\textbf{Unreachability.} One natural choice of deviation measure is the difficulty of reaching the baseline state $\base$ from the current state $\cur$. Reachability of the starting state $\start$ is commonly used as a constraint in safe exploration methods~\citep{Moldovan12,Eysenbach17}, where the agent does not take an action if it makes the reachability value function too low.

We define reachability of state $y$ from state $x$ as the value function of the optimal policy given a reward of 1 for reaching $y$ and 0 otherwise:
$$\reach{x}{y} := \max_\pi \expect \gammar^{\ns{\pi}{x}{y}}$$
where $\ns{\pi}{x}{y}$ is the number of steps it takes to reach $y$ from $x$ when following policy $\pi$, and $\gammar \in (0,1]$ is the reachability discount factor. This can be computed recursively as follows:
\begin{align*}
\reach{x}{y} &= \gammar \max_{a} \sum_{z \in \states} \trans(z | x, a) \reach{z}{y} \text{ for } x \not= y \\
\reach{y}{y} &= 1
\end{align*}

A special case is \emph{undiscounted} reachability ($\gammar=1$), which computes whether $y$ is reachable in any number of steps. We show that undiscounted reachability reduces to
$$\reach{x}{y}= \max_\pi P(\ns{\pi}{x}{y} < \infty). $$

The \emph{unreachability (UR)} deviation measure is then defined as
$$\dev{UR} := 1 - \reach{\cur}{\base}.$$

The undiscounted unreachability measure only penalizes irreversible transitions, while the discounted measure also penalizes reversible transitions. 

\begin{figure}[ht]
\centering
\includegraphics[scale=\sizeB]{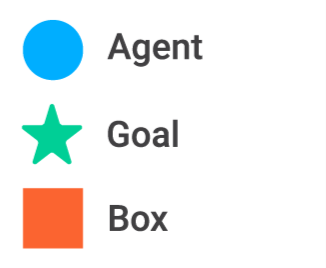}
\includegraphics[scale=\sizeB]{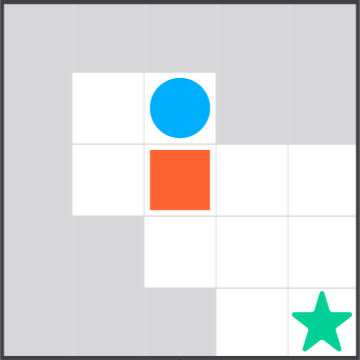}
\caption{Box environment.}
\label{fig:box_gridworld}
\end{figure}

A problem with the unreachability measure is that it takes the maximum value of 1 if the agent takes any irreversible action (since the reachability of the baseline becomes 0). Thus, the agent receives the maximum penalty independently of the magnitude of the irreversible action, e.g. whether the agent breaks one vase or a hundred vases. This can lead to unsafe behavior, as demonstrated on the Box environment from the AI Safety Gridworlds suite ~\citep{Leike17}, shown in Figure \ref{fig:box_gridworld}.

The environment contains a box that needs to be pushed out of the way for the agent to reach the goal. The unsafe behavior is taking the shortest path to the goal, which involves pushing the box down into a corner (an irrecoverable position). The desired behavior is to take a slightly longer path in order to push the box to the right. 

The action of moving the box is irreversible in both cases: if the box is moved to the right, the agent can move it back, but then the agent ends up on the other side of the box. Thus, the agent receives the maximum penalty of 1 for moving the box in any direction, so the penalty does not incentivize the agent to choose the safe path. Section \ref{sec:experiments} confirms that the unreachability penalty fails on the Box environment for all choices of baseline.

\textbf{Relative reachability.} To address the magnitude-sensitivity problem, we now introduce a reachability-based measure that is sensitive to the magnitude of the irreversible action. We define the \emph{relative reachability (RR) measure} as the average reduction in reachability of all states $s$ from the current state $\cur$ compared to the baseline $\base$:
$$\dev{RR} := \frac{1}{|\states|} \sum_{s\in\states} \tr{\reach{\base}{s} - \reach{\cur}{s}}$$

The RR measure is nonnegative everywhere, and zero for states $\cur$ that reach or exceed baseline reachability of all states. See Figure \ref{fig:rr} for an illustration. 

In the Box environment, moving the box down makes more states unreachable than moving the box to the right (in particular, all states where the box is not in a corner become unreachable). Thus, the agent receives a higher penalty for moving the box down, and has an incentive to move the box to the right.

\ifarxiv
\textbf{Attainable utility}
Another magnitude-sensitive deviation measure, which builds on the presentation of the RR measure in the first version of this paper, is the attainable utility (AU) measure \citep{Turner19}.
Observing that the informal notion of value may be richer than mere reachability of states, AU considers a set $\rewards$ of arbitrary reward functions.
We can define this measure as follows:
\begin{align*}
\dev{AU} &:= \frac{1}{|\rewards|}\sum_{r\in{\rewards}}|V_r(s_t')-V_r(s_t)|\\
\text{ where } V_r(\tilde{s}) &:= \max_\pi\sum_{t=0}^\infty\gamma_r^k{x(\tilde s_t^\pi)}
\end{align*}
is the value of state $\tilde{s}$ according to reward function $r$ (here $\tilde s_t^\pi$ denotes the state obtained from $\tilde s$ by following $\pi$ for $t$ steps).

In the Box environment, the AU measure gives a higher penalty for moving the box into a corner, since this affects the attainability of reward functions that reward states where the box is not in the corner. Thus, similarly to the RR measure, it incentivizes the agent to move the box to the right.

\textbf{Value-difference measures.}
The RR and AU deviation measures are examples of what we call \emph{value-difference measures}, whose general form is:
$$\dev{VD} := \sum_x w_x f(V_x(s_t') - V_x(s_t))$$
where $x$ ranges over some sources of value, $V_x(\tilde{s})$ is the value of state $\tilde{s}$ according to $x$, $w_x$ is a weighting or normalizing factor, and $f$ is the function for summarizing the value difference.
Thus value-difference measures calculate a weighted summary of the differences in measures of value between the current and baseline states.

For RR, we take $x$ to range over states in $S$ and $V_x(\tilde{s}) = R(\tilde{s}, x)$, so the sources of value are, for each state, the reachability of that state. We take $w_x=1/|S|$ and $f(d) = \max(d, 0)$ (``truncated difference''), which penalizes decreases (but not increases) in value.
For AU, we take $x$ to range over reward functions in $\rewards$ and  $V_x(\tilde{s})$ as above, so the sources of value are, for each reward function, the maximum attainable reward according to that function. We take  $w_x=1/|\rewards|$ and $f(d)=|d|$ (``absolute difference''), which penalizes all changes in value.
The choice of \emph{summary function} $f$ is orthogonal to the other choices: we can also consider absolute difference for RR and truncated difference for AU.

One can view AU as a generalization of RR under certain conditions: namely, if we have one reward function per state that assigns value $1$ to that state and $0$ otherwise, assuming the state cannot be reached again later in the same episode.

\textbf{Modifications required with the stepwise inaction baseline.} In order to capture the delayed effects of actions, we modify each of the deviation measures to incorporate the inaction rollouts from $\cur$ and $\base=\step$ (shown in Figure \ref{fig:rollouts}).
We denote the modified measure with an $S$ in front (for `stepwise inaction baseline').
\begin{align*}
\dev{SUR} :=& 1 - (1-\gamma)\sum_{k=0}^{\infty} \gamma^k \reach{\rollout{s}{t}{t+k}}{\rollout{s'}{t}{t+k}}\\
\dev{SVD} :=& \sum_{x} w_{x}f({RV_{x}(\base) - RV_{x}(\cur)})\\
\text{where } RV_x(\tilde \cur) :=& (1-\gamma) \sum_{k=0}^{\infty} \gamma^k V_x(\rollout{\tilde s}{t}{t+k})
\end{align*}
We call $RV_{x}(\tilde \cur)$ the \emph{rollout value} of $\tilde \cur \in \{\cur, \base\}$ with respect to $x$. 
In a deterministic environment, the UR measure $\dev{SUR}$ and the rollout value $RV_{x}(\tilde \cur)$ used in the value difference measures $\dev{SRR}$ and $\dev{SAU}$ can be computed recursively as follows:
\begin{align*}
\devx{SUR}{s_1}{s_2} =& (1-\gamma) (\reach{s_1}{s_2} + \gamma \devx{SUR}{I(s_1)}{I(s_2)})\\
RV_{x}(s) =& (1-\gamma) (V_{x}(s) + \gamma RV_{x}(I(s)))
\end{align*}
where $I(s)$ is the inaction function that gives the state reached by following the inaction policy from state $s$ (this is the identity function in static environments).
\else
\textbf{Modifications for the stepwise inaction baseline.} In order to capture the delayed effects of actions, we modify the deviation measures to incorporate the inaction rollouts from $\cur$ and $\base=\step$ (shown in Figure \ref{fig:rollouts}) as follows:
\begin{align*}
\dev{SUR} :=& 1 - (1-\gamma)\sum_{k=0}^{\infty} \gamma^k \reach{\rollout{s}{t}{t+k}}{\rollout{s'}{t}{t+k}}\\
\rv{\tilde \cur}{s} :=& (1-\gamma) \sum_{k=0}^{\infty} \gamma^k \reach{\rollout{\tilde s}{t}{t+k}}{s} \\
\dev{SRR} :=& \frac{1}{|\states|} \sum_{s\in\states} \tr{\rv{\base}{s} - \rv{\cur}{s}} 
\end{align*}
We call $\rv{\tilde \cur}{s}$ the \emph{rollout value} of $\tilde \cur \in \{\cur, \base\}$ with respect to $s$. 
In a deterministic environment, the UR measure $\dev{SUR}$ and the rollout value $\rv{\tilde \cur}{s}$ for the RR measure $\dev{SRR}$ can be computed recursively as follows:
\begin{align*}
\devx{SUR}{x}{y} =& (1-\gamma) (\reach{x}{y} + \gamma \devx{SUR}{I(x)}{I(y)})\\
\rv{x}{y} =& (1-\gamma) (\reach{x}{y} + \gamma \rv{I(x)}{y})
\end{align*}
where $I(x)$ is the inaction function that gives the state reached by following the inaction policy from state $x$ (this is the identity function in static environments).
\fi

\ifarxiv
\section{Experiments}\label{sec:experiments}
\else
\section{EXPERIMENTS}\label{sec:experiments}
\fi

We run a tabular Q-learning agent with different penalties on the gridworld environments introduced in Section \ref{sec:components}. While these environments are simplistic, they provide a proof of concept by clearly illustrating the desirable and undesirable behaviors, which would be more difficult to isolate in more complex environments. 
We compare all combinations of the following design choices for an impact penalty:
\begin{itemize}
    \item Baselines: starting state $\start$, inaction $\inaction$, stepwise inaction $\step$
\ifarxiv
    \item Deviation measures: unreachability (UR) ($\dev{SUR}$ for the stepwise inaction baseline, $\dev{UR}$ for other baselines), and the value-difference measures, relative reachability (RR) and attainable utility (AU) ($\dev{SVD}$ for the stepwise inaction baseline, $\dev{VD}$ for the other baselines, for $VD\in\{RR,AU\}$).
    \item Discounting: $\gammar = 0.99$ (discounted), $\gammar=1.0$ (undiscounted). (We omit the undiscounted case for AU due to convergence issues.)
    \item Summary functions: truncation $f(d) = \max(d, 0)$, absolute $f(d) = |d|$
\else
    \item Deviation measures: unreachability (UR) ($\dev{SUR}$ for the stepwise inaction baseline $\dev{UR}$ for other baselines), relative reachability (RR) ($\dev{SRR}$ for the stepwise inaction baseline, $\dev{RR}$ for other baselines)
    \item Discounting: $\gammar = 0.99$ (discounted), $\gammar=1.0$ (undiscounted)
\fi
\end{itemize}

\ifarxiv\else
We also compare to no penalty, shown as None in Figure \ref{fig:results}.
\fi

The reachability function $R$ is approximated based on states and transitions that the agent has encountered. It is initialized as $\reach{x}{y} = 1$ if $x=y$ and 0 otherwise (different states are unreachable from each other). When the agent makes a transition $(\cur, \act, \nxt)$, we make a shortest path update to the reachability function. For any two states  $x$ and $y$ where $\cur$ is reachable from $x$, and $y$ is reachable from $\nxt$, we update $\reach{x}{y}$. This approximation assumes a deterministic environment. 

\ifarxiv
Similarly, the value functions $V_r$ used for attainable utility are approximated based on the states and transitions encountered.
For each state $y$, we track the set of states $x$ for which a transition to $y$ has been observed. When the agent makes a transition, we make a Bellman update to the value function of each reward function $r$, setting $V_r(x)\leftarrow\max(V_r(x), u(x) + \gammar{V_r(y)})$ for all pairs of states such that $y$ is known to be reachable from $x$ in one step.
\else\fi

We use a perfect environment model to obtain the outcomes of noop actions $a^\noop$ for the inaction and stepwise inaction baselines. We leave model-free computation of the baseline to future work.

In addition to the reward function, each environment has a \emph{performance} function, originally introduced by~\citet{Leike17}, which is not observed by the agent. This represents the agent's performance according to the designer's true preferences: it reflects how well the agent achieves the objective and whether it does so safely.

We anneal the exploration rate linearly from 1 to 0 over 9000 episodes, and keep it at 0 for the next 1000 episodes. For each penalty on each environment, we use a grid search to tune the scaling parameter $\beta$, choosing the value of $\beta$ that gives the highest average performance on the last 100 episodes.
(The grid search is over $\beta=0.1, 0.3, 1, 3, 10, 30, 100, 300$.)
Figure \ref{fig:results} shows scaled performance results for all penalties, where a value of 1 corresponds to optimal performance (achieved by the desired behavior), and a value of 0 corresponds to undesired behavior (such as interference or offsetting). 

\ifarxiv

\begin{figure}[hp!]
\begin{subfigure}{\textwidth}
\centering
\includegraphics[scale=0.6]{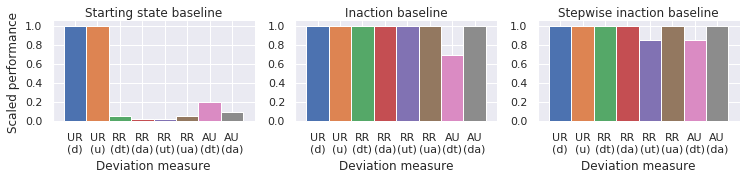}
\caption{Sushi environment}\label{fig:sushi_results}
\end{subfigure}
\begin{subfigure}{\textwidth}
\centering
\includegraphics[scale=0.6]{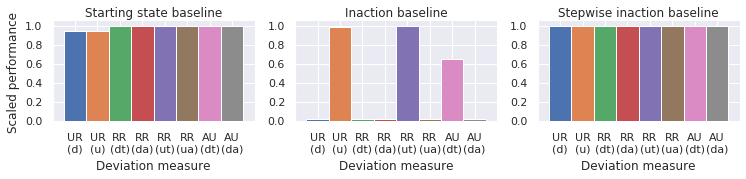}
\caption{Vase environment}\label{fig:vase_results}
\end{subfigure}
\begin{subfigure}{\textwidth}
\centering
\includegraphics[scale=0.6]{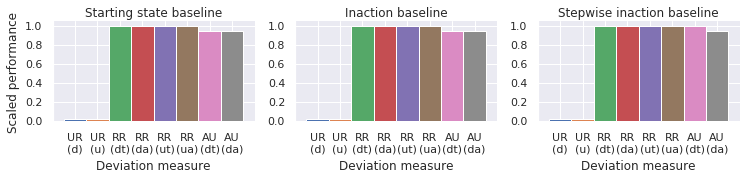}
\caption{Box environment}\label{fig:box_results}
\end{subfigure}
\begin{subfigure}{\textwidth}
\centering
\includegraphics[scale=0.6]{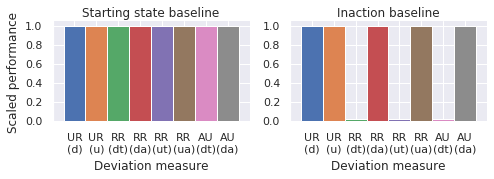}
\caption{Survival environment}\label{fig:survival_results}
\end{subfigure}

\caption{Scaled performance results for different combinations of design choices (averaged over 20 seeds). The columns are different baseline choices: starting state, inaction, and stepwise inaction. The bars in each plot are results for different deviation measures (UR, RR and AU), with discounted and undiscounted versions indicated by (d) and (u) respectively, and truncation and absolute functions indicated by (t) and (a) respectively. 1 is optimal performance and 0 is the performance achieved by unsafe behavior (when the box is pushed into a corner, the vase is broken, the sushi is taken off the belt, or the off switch is disabled).}
\label{fig:results}
\end{figure}

\else

\begin{figure*}[ht]
\centering
\includegraphics[scale=0.6]{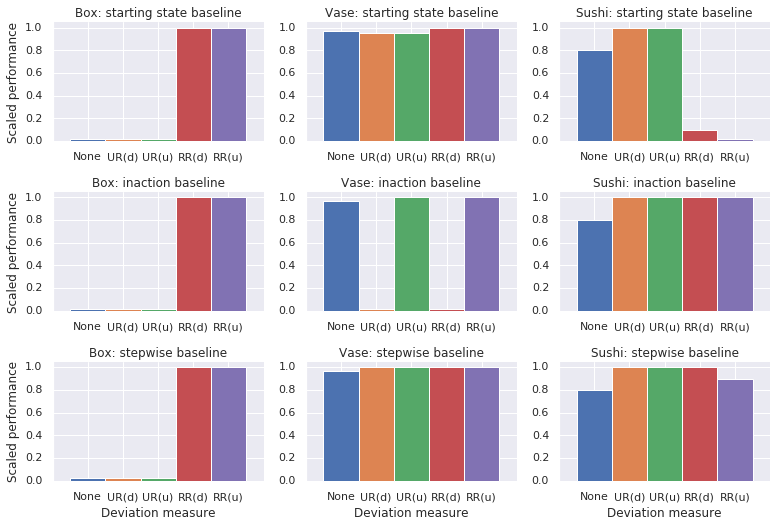}
\caption{Scaled performance results for different combinations of design choices (averaged over 10 seeds). The columns are the Box, Vase and Sushi environments, and the rows are different baseline choices (starting state, inaction, and stepwise inaction). The bars in each plot are results for different deviation measures (None, UR and RR), with discounted and undiscounted versions indicated by (d) and (u) respectively. 1 is optimal performance and 0 is the performance achieved by unsafe behavior (when the sushi is taken off the belt, the vase is broken, or the box is pushed into a corner).}
\label{fig:results}
\end{figure*}
\fi

\textbf{Sushi environment.} The environment is shown in Figure \ref{fig:sushi_gridworld}.
The agent receives a reward of 50 for reaching the goal (which terminates the episode), and no movement penalty. An agent with no penalty achieves scaled performance 0.8 (avoiding interference most of the time).
Here, all penalties with the inaction and stepwise inaction baselines reach near-optimal performance. 
\ifarxiv
The RR and AU penalties with the starting state baseline produce
\else
The RR penalty with the starting state baseline produces 
\fi
the interference behavior (removing the sushi from the belt), resulting in scaled performance 0. However, since the starting state is unreachable no matter what the agent does, the UR penalty is always at the maximum value of 1, so similarly to no penalty, it does not produce interference behavior. 
\ifarxiv
The discounting and summary function choices don't make much difference on this environment.
\fi

\ifarxiv
\begin{table}[ht]
\centering
\begin{tabular}{c||c|c|c}
& \textcolor{red}{Starting state} & Inaction & Stepwise inaction \\\hline
UR   & \checkmark    & \checkmark    & \checkmark  \\
RR   & \xmark        & \checkmark    & \checkmark  \\
AU   & \xmark        & \checkmark    & \checkmark                
\end{tabular}
\caption{Sushi environment summary.}\label{tab:sushi_results}
\end{table}
\else
\begin{table}[ht]
\centering
\caption{Sushi environment summary.}\label{tab:box_results}
\begin{tabular}{c||c|c|c}
& \textcolor{red}{Starting} & Inaction & Stepwise \\\hline
None  & \checkmark    & \checkmark    & \checkmark   \\
UR    & \checkmark    & \checkmark    & \checkmark   \\
RR    & \xmark        & \checkmark    & \checkmark   \\ 
\end{tabular}
\end{table}
\fi

\textbf{Vase environment.}
The environment is shown in Figure \ref{fig:vase_gridworld}. The agent receives a reward of 50 for taking the vase off the belt. The episode lasts 20 steps, and there is no movement penalty. An agent with no penalty achieves scaled performance 0.98.
Unsurprisingly, all penalties with the starting state baseline perform well here. 
\ifarxiv
With the inaction baseline, the discounted UR and RR penalties receive scaled performance 0, which corresponds to the offsetting behavior of moving the vase off the belt and then putting it back on, shown in Figure \ref{fig:offsetting}. Surprisingly, discounted AU with truncation avoids offsetting some of the time, which is probably due to convergence issues. The undiscounted versions with the truncation function avoid this behavior, since the action of taking the vase off the belt is reversible and thus is not penalized at all, so there is nothing to offset. All penalties with the absolute function produce the offsetting behavior, since removing the vase from the belt is always penalized.
\else
With the inaction baseline, the discounted penalties receive scaled performance 0, which corresponds to the offsetting behavior of moving the vase off the belt and then putting it back on, shown in Figure \ref{fig:offsetting}. The undiscounted versions avoid this behavior, since the action of taking the vase off the belt is reversible and thus is not penalized at all, so there is nothing to offset. 
\fi
All penalties with the stepwise inaction baseline perform well on this environment, showing that this baseline does not produce the offsetting incentive.

\ifarxiv
\begin{table}[ht]
\centering
\begin{tabular}{c||c|c|c}
& Starting state & \textcolor{red}{Inaction} & Stepwise inaction \\\hline
UR (discounted)               & \checkmark  & \xmark      & \checkmark \\
UR (undiscounted)             & \checkmark  & \checkmark  & \checkmark \\
RR (discounted, truncation)   & \checkmark  & \xmark      & \checkmark \\
RR (discounted, absolute)     & \checkmark  & \xmark      & \checkmark \\
RR (undiscounted, truncation) & \checkmark  & \checkmark  & \checkmark \\
RR (undiscounted, absolute)   & \checkmark  & \xmark      & \checkmark \\
AU (discounted, truncation)   & \checkmark  & ?           & \checkmark \\
AU (discounted, absolute)     & \checkmark  & \xmark      & \checkmark 
\end{tabular}
\caption{Vase environment summary.}\label{tab:vase_results}
\end{table}
\else
\begin{table}[ht]
\centering
\caption{Vase environment summary.}\label{tab:vase_results}
\begin{tabular}{c||c|c|c}
& Starting  & \textcolor{red}{Inaction} & Stepwise  \\\hline
None               & \checkmark  & \checkmark  & \checkmark \\
UR (discounted)    & \checkmark  & \xmark      & \checkmark \\
UR (undiscounted)  & \checkmark  & \checkmark  & \checkmark \\
RR (discounted)    & \checkmark  & \xmark      & \checkmark \\
RR (undiscounted)  & \checkmark  & \checkmark  & \checkmark \\
\end{tabular}
\end{table}
\fi

\textbf{Box environment.}
The environment is shown in Figure \ref{fig:box_gridworld}. The agent receives a reward of 50 for reaching the goal (which terminates the episode), and a movement penalty of -1. The starting state and inaction baseline are the same in this environment, while the stepwise inaction baseline is different. 
The safe longer path to the goal receives scaled performance 1, while the unsafe shorter path that puts the box in the corner receives scaled performance 0. An agent with no penalty achieves scaled performance 0.
\ifarxiv
For all baselines, RR and AU achieve near-optimal scaled performance, while UR achieves scaled performance 0. 
\else
For all baselines, RR achieves optimal scaled performance 1, while UR achieves scaled performance 0. 
\fi
This happens because the UR measure is not magnitude-sensitive, and thus does not distinguish between irreversible actions that result in recoverable and irrecoverable box positions, as described in Section \ref{sec:deviation}. 

\ifarxiv
\begin{table}[ht]
\centering
\begin{tabular}{c||c|c|c}
& Starting state & Inaction & Stepwise inaction \\\hline
\textcolor{red}{UR} & \xmark      & \xmark      & \xmark     \\
RR                  & \checkmark  & \checkmark  & \checkmark \\ 
AU                  & \checkmark  & \checkmark  & \checkmark
\end{tabular}
\caption{Box environment summary.}\label{tab:box_results}
\end{table}
\else
\begin{table}[ht]
\centering
\caption{Box environment summary.}\label{tab:box_results}
\begin{tabular}{c||c|c|c}
& Starting & Inaction & Stepwise \\\hline
\textcolor{red}{None} & \xmark      & \xmark      & \xmark     \\
\textcolor{red}{UR}   & \xmark      & \xmark      & \xmark     \\
RR                    & \checkmark  & \checkmark  & \checkmark \\ 
\end{tabular}
\end{table}
\fi


\ifarxiv
Overall, the combinations of design choices that perform best across all environments are RR and AU with the stepwise inaction baseline and undiscounted RR with the inaction baseline.
\else
Overall, the combinations of design choices that perform best across all environments are RR with the stepwise inaction baseline and undiscounted RR with the inaction baseline. 
\fi
Since the undiscounted RR measure only penalizes irreversible transitions, a penalty that aims to penalize reversible transitions as well has to be combined with the stepwise inaction baseline.

\ifarxiv
\textbf{Effects on interruptibility.} We also examine the effects of impact measure design choices on whether the agent is interruptible~\citep{Orseau16}, using the Survival Incentive environment introduced in \cite{Turner19}. In this environment, the agent has the option to disable an off switch, which prevents the episode from ending before the agent reaches the goal. We only have results for the inaction baseline, since the deterministic assumption for our implementation of recursive inaction rollouts doesn't hold in this environment. Results for the stepwise inaction baseline with non-recursive rollouts are given in~\cite{Turner19}.

All penalties with the starting state baseline avoid disabling the off switch, probably because this is an irreversible action. With the inaction baseline, the decrease-penalizing variants of RR and AU disable the off switch, while the difference-penalizing variants do not, as shown in Figure \ref{fig:survival_results}. (Note that this does not hold in the Safe Interruptibility environment in the AI Safety Gridworlds suite, where interruption is implemented as the agent getting stuck rather than terminating the episode.) However, penalizing differences in reachability or attainable utility also has downsides, since this can impede the agent's ability to create desirable change in the environment more than penalizing decreases.

\begin{table}[ht]
\centering
\begin{tabular}{c||c|c}
& Starting state & \textcolor{red}{Inaction} \\\hline
UR (discounted)               & \checkmark  & \checkmark \\
UR (undiscounted)             & \checkmark  & \checkmark \\
RR (discounted, \textcolor{red}{truncation})   & \checkmark  & \xmark     \\
RR (discounted, absolute)     & \checkmark  & \checkmark \\
RR (undiscounted, \textcolor{red}{truncation}) & \checkmark  & \xmark     \\
RR (undiscounted, absolute)   & \checkmark  & \checkmark \\
AU (discounted, \textcolor{red}{truncation})   & \checkmark  & \xmark     \\
AU (discounted, absolute)     & \checkmark  & \checkmark 
\end{tabular}
\caption{Vase environment summary.}\label{tab:survival_results}
\end{table}

\fi

\ifarxiv
\section{Additional related work}
\else
\section{ADDITIONAL RELATED WORK}
\fi

\textbf{Safe exploration.} Safe exploration methods prevent the agent from taking harmful actions by enforcing safety constraints~\citep{Turchetta16, Dalal18}, penalizing risk~\citep{Chow15,Mihatsch02}, using intrinsic motivation~\citep{Lipton16}, preserving reversibility~\citep{Moldovan12, Eysenbach17}, etc. Explicitly defined constraints or safe regions tend to be task-specific and require significant human input, so they do not provide a general solution to the side effects problem. Penalizing risk and intrinsic motivation can help the agent avoid low-reward states (such as getting trapped or damaged), but do not discourage the agent from damaging the environment if this is not accounted for in the reward function. Reversibility-preserving methods produce interference and magnitude insensitivity incentives as discussed in Section \ref{sec:components}.

\textbf{Side effects criteria using state features.} \cite{Zhang18} assumes a factored MDP where the agent is allowed to change some of the features and proposes a criterion for querying the supervisor about changing other features in order to allow for intended effects on the environment. \cite{Shah19} define an auxiliary reward for avoiding side effects in terms of state features by assuming that the starting state of the environment is already organized according to human preferences. Since the latter method uses the starting state as a baseline, we would expect it to produce interference behavior in dynamic environments. While these approaches are promising, they are not general in their present form due to reliance on state features.

\textbf{Empowerment.}
Our RR measure is related to \emph{empowerment} \citep{Klyubin05,Salge14,Mohamed15,Gregor17}, a measure of the agent's control over its environment, defined as the highest possible mutual information between the agent's actions and the future state. Empowerment measures the agent's ability to reliably reach many states, while RR penalizes the reduction in reachability of states relative to the baseline. Maximizing empowerment would encourage the agent to avoid irreversible side effects, but would also incentivize interference behavior, and it is unclear to us how to define an empowerment-based measure that would avoid this. One possibility would be to penalize the reduction in empowerment between the current state $\cur$ and the baseline $\base$. However, empowerment is indifferent between these two situations: A) the same states are reachable from $\cur$ and $\base$, and B) a state $s_1$ reachable from $\base$ but unreachable from $\cur$, while another state $s_2$ is reachable from $\cur$ but unreachable from $\base$. Thus, penalizing reduction in empowerment would miss some side effects: e.g. if the agent replaced the sushi on the conveyor belt with a vase, empowerment could remain the same, and so the agent would not be penalized for destroying the vase.

\textbf{Uncertainty about the objective.} Inverse Reward Design~\citep{Hadfield-Menell17} incorporates uncertainty about the objective by considering alternative reward functions that are consistent with the given reward function in the training environment. This helps the agent avoid some side effects that stem from distributional shift, where the agent encounters a new state that was not present in training. However, this method assumes that the given reward function is correct for the training environment, and so does not prevent side effects caused by a reward function that is misspecified in the training environment.
Quantilization~\citep{Taylor2016quant} incorporates uncertainty by taking actions from the top quantile of actions, rather than the optimal action. These methods help to prevent side effects, but do not provide a way to quantify side effects.


\textbf{Human oversight.} An alternative to specifying a side effects penalty is to teach the agent to avoid side effects through human oversight, such as inverse reinforcement learning~\citep{Ng00, Ziebart08, Hadfield-Menell16}, demonstrations~\citep{Abbeel04,Hester18}, or human feedback~\citep{Christiano17, Saunders17, Warnell18}. It is unclear how well an agent can learn a general heuristic for avoiding side effects from human oversight. We expect this to depend on the diversity of settings in which it receives oversight and its ability to generalize from those settings, which are difficult to quantify. 
We expect that an intrinsic penalty for side effects would be more robust and more reliably result in avoiding them. Such a penalty could also be combined with human oversight to decrease the amount of human input required for an agent to learn human preferences.

\ifarxiv
\section{Conclusions}
\else
\section{CONCLUSIONS}
\fi

We have outlined a set of bad incentives (interference, offsetting, and magnitude insensitivity) that can arise from a poor choice of baseline or deviation measure, and proposed design choices that avoid these incentives in preliminary experiments. 
There are many possible directions where we would like to see follow-up work:



\textbf{Scalable implementation.} The RR measure in its exact form is not tractable for environments more complex than gridworlds. In particular, we compute reachability between all pairs of states, and use an environment simulator to compute the baseline. 
A more practical implementation could be computed over some set of representative states instead of all states. For example, the agent could learn a set of auxiliary policies for reaching distinct states, similarly to the method for approximating empowerment in \cite{Gregor17}. 


\textbf{Better choices of baseline.} 
Using noop actions to define inaction for the stepwise inaction baseline can be problematic, since the agent is not penalized for causing side effects that would occur in the noop baseline. For example, if the agent is driving a car on a winding road, then at any point the default outcome of a noop is a crash, so the agent would not be penalized for spilling coffee in the car. This could be avoided using a better inaction baseline, such as following the road, but this can be challenging to define in a task-independent way.

\textbf{Theory.} There is a need for theoretical work on characterizing and formalizing  undesirable incentives that arise from different design choices in penalizing side effects.

\textbf{Taking into account reward costs.} While the discounted relative reachability measure takes into account the time costs of reaching various states, it does not take into account reward costs. For example, suppose the agent can reach state $s$ from the current state in one step, but this step would incur a large negative reward. Discounted reachability could be modified to reflect this by adding a term for reward costs.

\textbf{Weights over the state space.} In practice, we often value the reachability of some states much more than others. This could be incorporated into the relative reachability measure by adding a weight $w_s$ for each state $s$ in the sum. Such weights could be learned through human feedback methods, e.g. \cite{Christiano17}.

We hope this work lays the foundations for a practical methodology on avoiding side effects that would scale well to more complex environments.

\ifarxiv

\section*{Acknowledgements}

We are grateful to Jan Leike, Pedro Ortega, Tom Everitt, Alexander Turner, David Krueger, Murray Shanahan, Janos Kramar, Jonathan Uesato, Tam Masterson and Owain Evans for helpful feedback on drafts. We would like to thank them and Toby Ord, Stuart Armstrong, Geoffrey Irving, Anthony Aguirre, Max Wainwright, Jaime Fisac, Rohin Shah, Jessica Taylor, Ivo Danihelka, and Shakir Mohamed for illuminating conversations.

\else

\renewcommand{\bibsection}{\subsubsection*{References}}

\fi

\bibliography{main}

\begin{thebibliography}{36}
\providecommand{\natexlab}[1]{#1}
\providecommand{\url}[1]{\texttt{#1}}
\expandafter\ifx\csname urlstyle\endcsname\relax
  \providecommand{\doi}[1]{doi: #1}\else
  \providecommand{\doi}{doi: \begingroup \urlstyle{rm}\Url}\fi

\bibitem[Abbeel and Ng(2004)]{Abbeel04}
Pieter Abbeel and Andrew Ng.
\newblock Apprenticeship learning via inverse reinforcement learning.
\newblock In \emph{International Conference on Machine Learning}, pages 1--8,
  2004.

\bibitem[Amodei et~al.(2016)Amodei, Olah, Steinhardt, Christiano, Schulman, and
  Mané]{AmodeiOlah16}
Dario Amodei, Chris Olah, Jacob Steinhardt, Paul Christiano, John Schulman, and
  Dan Mané.
\newblock Concrete problems in {AI} safety.
\newblock \emph{arXiv preprint arXiv:1606.06565}, 2016.

\bibitem[Armstrong and Levinstein(2017)]{Armstrong17}
Stuart Armstrong and Benjamin Levinstein.
\newblock Low impact artificial intelligences.
\newblock \emph{arXiv preprint arXiv:1705.10720}, 2017.

\bibitem[Chow et~al.(2015)Chow, Tamar, Mannor, and Pavone]{Chow15}
Yinlam Chow, Aviv Tamar, Shie Mannor, and Marco Pavone.
\newblock Risk-sensitive and robust decision-making: a {CVaR} optimization
  approach.
\newblock In \emph{Neural Information Processing Systems (NIPS)}, pages
  1522--1530, 2015.

\bibitem[Christiano et~al.(2017)Christiano, Leike, Brown, Martic, Legg, and
  Amodei]{Christiano17}
Paul Christiano, Jan Leike, Tom~B Brown, Miljan Martic, Shane Legg, and Dario
  Amodei.
\newblock Deep reinforcement learning from human preferences.
\newblock In \emph{Neural Information Processing Systems (NIPS)}, 2017.

\bibitem[Dalal et~al.(2018)Dalal, Dvijotham, Vecerik, Hester, Paduraru, and
  Tassa]{Dalal18}
Gal Dalal, Krishnamurthy Dvijotham, Matej Vecerik, Todd Hester, Cosmin
  Paduraru, and Yuval Tassa.
\newblock Safe exploration in continuous action spaces.
\newblock \emph{arXiv preprint arXiv:1801.08757}, 2018.

\bibitem[Eysenbach et~al.(2017)Eysenbach, Gu, Ibarz, and Levine]{Eysenbach17}
Benjamin Eysenbach, Shixiang Gu, Julian Ibarz, and Sergey Levine.
\newblock Leave no {T}race: Learning to reset for safe and autonomous
  reinforcement learning.
\newblock \emph{arXiv preprint arXiv:1711.06782}, 2017.

\bibitem[Fisac et~al.(2017)Fisac, Akametalu, Zeilinger, Kaynama, Gillula, and
  Tomlin]{Fisac17}
Jaime~F. Fisac, Anayo~K. Akametalu, Melanie~Nicole Zeilinger, Shahab Kaynama,
  Jeremy~H. Gillula, and Claire~J. Tomlin.
\newblock A general safety framework for learning-based control in uncertain
  robotic systems.
\newblock \emph{arXiv preprint arXiv:1705.01292}, 2017.

\bibitem[García and Fernández(2015)]{Garcia15}
Javier García and Fernando Fernández.
\newblock A comprehensive survey on safe reinforcement learning.
\newblock \emph{Journal of Machine Learning Research}, 16\penalty0
  (1):\penalty0 1437--1480, 2015.

\bibitem[Gillula and Tomlin(2012)]{Gillula12}
Jeremy~H. Gillula and Claire~J. Tomlin.
\newblock Guaranteed safe online learning via reachability: tracking a ground
  target using a quadrotor.
\newblock In \emph{{IEEE} International Conference on Robotics and Automation
  (ICRA)}, pages 2723--2730, 2012.

\bibitem[Gregor et~al.(2017)Gregor, Rezende, and Wierstra]{Gregor17}
Karol Gregor, Danilo~Jimenez Rezende, and Daan Wierstra.
\newblock Variational intrinsic control.
\newblock \emph{International Conference for Learning Representations (ICLR)
  Workshop, arXiv preprint arXiv:1611.07507}, 2017.

\bibitem[Hadfield-Menell et~al.(2016)Hadfield-Menell, Dragan, Abbeel, and
  Russell]{Hadfield-Menell16}
Dylan Hadfield-Menell, Anca Dragan, Pieter Abbeel, and Stuart Russell.
\newblock Cooperative inverse reinforcement learning.
\newblock In \emph{Neural Information Processing Systems (NIPS)}, 2016.

\bibitem[Hadfield{-}Menell et~al.(2017)Hadfield{-}Menell, Milli, Abbeel,
  Russell, and Dragan]{Hadfield-Menell17}
Dylan Hadfield{-}Menell, Smitha Milli, Pieter Abbeel, Stuart~J. Russell, and
  Anca~D. Dragan.
\newblock Inverse reward design.
\newblock In \emph{Neural Information Processing Systems (NIPS)}, pages
  6768--6777, 2017.

\bibitem[Hester et~al.(2018)Hester, Vecerik, Pietquin, Lanctot, Schaul, Piot,
  Horgan, Quan, Sendonaris, Osband, Dulac{-}Arnold, Agapiou, Leibo, and
  Gruslys]{Hester18}
Todd Hester, Matej Vecerik, Olivier Pietquin, Marc Lanctot, Tom Schaul, Bilal
  Piot, Dan Horgan, John Quan, Andrew Sendonaris, Ian Osband, Gabriel
  Dulac{-}Arnold, John Agapiou, Joel~Z. Leibo, and Audrunas Gruslys.
\newblock Deep q-learning from demonstrations.
\newblock In \emph{{AAAI} Conference on Artificial Intelligence}, 2018.

\bibitem[Klyubin et~al.(2005)Klyubin, Polani, and Nehaniv]{Klyubin05}
Alexander~S. Klyubin, Daniel Polani, and Chrystopher~L. Nehaniv.
\newblock All else being equal be empowered.
\newblock In \emph{European Conference on Artificial Life (ECAL)}, pages
  744--753, 2005.

\bibitem[Leike et~al.(2017)Leike, Martic, Krakovna, Ortega, Everitt, Lefrancq,
  Orseau, and Legg]{Leike17}
Jan Leike, Miljan Martic, Victoria Krakovna, Pedro~A. Ortega, Tom Everitt,
  Andrew Lefrancq, Laurent Orseau, and Shane Legg.
\newblock {AI} safety gridworlds.
\newblock \emph{arXiv preprint arXiv:1711.09883}, 2017.

\bibitem[Lipton et~al.(2016)Lipton, Gao, Li, Chen, and Deng]{Lipton16}
Zachary~C. Lipton, Jianfeng Gao, Lihong Li, Jianshu Chen, and Li~Deng.
\newblock Combating reinforcement learning's sisyphean curse with intrinsic
  fear.
\newblock \emph{arXiv preprint arXiv:1611.01211}, 2016.

\bibitem[McCarthy and Hayes(1969)]{Mccarthy69}
John McCarthy and Patrick~J. Hayes.
\newblock Some philosophical problems from the standpoint of artificial
  intelligence.
\newblock In \emph{Machine Intelligence}, pages 463--502. Edinburgh University
  Press, 1969.

\bibitem[Mihatsch and Neuneier(2002)]{Mihatsch02}
Oliver Mihatsch and Ralph Neuneier.
\newblock Risk-sensitive reinforcement learning.
\newblock \emph{Machine Learning}, 49\penalty0 (2):\penalty0 267--290, 2002.

\bibitem[Mitchell et~al.(2005)Mitchell, Bayen, and Tomlin]{Mitchell05}
Ian~M. Mitchell, Alexandre~M. Bayen, and Claire~J. Tomlin.
\newblock A time-dependent hamilton-jacobi formulation of reachable sets for
  continuous dynamic games.
\newblock \emph{{IEEE} Transactions on Automatic Control}, 50\penalty0
  (7):\penalty0 947--957, 2005.

\bibitem[Mohamed and Rezende(2015)]{Mohamed15}
Shakir Mohamed and Danilo~J. Rezende.
\newblock Variational information maximisation for intrinsically motivated
  reinforcement learning.
\newblock In \emph{Neural Information Processing Systems (NIPS)}, pages
  2125--2133, 2015.

\bibitem[Moldovan and Abbeel(2012)]{Moldovan12}
Teodor~Mihai Moldovan and Pieter Abbeel.
\newblock Safe exploration in {M}arkov decision processes.
\newblock In \emph{International Conference on Machine Learning (ICML)}, pages
  1451--1458, 2012.

\bibitem[Ng and Russell(2000)]{Ng00}
Andrew Ng and Stuart Russell.
\newblock Algorithms for inverse reinforcement learning.
\newblock In \emph{International Conference on Machine Learning}, pages
  663--670, 2000.

\bibitem[Orseau and Armstrong(2016)]{Orseau16}
Laurent Orseau and Stuart Armstrong.
\newblock Safely interruptible agents.
\newblock In \emph{Uncertainty in Artificial Intelligence}, pages 557--566,
  2016.

\bibitem[Ortega et~al.(2018)Ortega, Maini, and et~al]{Ortega18}
Pedro Ortega, Vishal Maini, and et~al.
\newblock Building safe artificial intelligence: specification, robustness, and
  assurance.
\newblock DeepMind Safety Research Blog, 2018.

\bibitem[Pecka and Svoboda(2014)]{Pecka14}
Martin Pecka and Tomas Svoboda.
\newblock Safe exploration techniques for reinforcement learning --- an
  overview.
\newblock In \emph{International Workshop on Modelling and Simulation for
  Autonomous Systems}, pages 357--375, 2014.

\bibitem[Salge et~al.(2014)Salge, Glackin, and Polani]{Salge14}
Christoph Salge, Cornelius Glackin, and Daniel Polani.
\newblock Empowerment --- an introduction.
\newblock In \emph{Guided Self-Organization: Inception}, pages 67--114.
  Springer, 2014.

\bibitem[Saunders et~al.(2017)Saunders, Sastry, Stuhlmueller, and
  Evans]{Saunders17}
William Saunders, Girish Sastry, Andreas Stuhlmueller, and Owain Evans.
\newblock Trial without error: Towards safe reinforcement learning via human
  intervention.
\newblock \emph{arXiv preprint arXiv:1707.05173}, 2017.

\bibitem[Shah et~al.(2019)Shah, Krasheninnikov, Alexander, Abbeel, and
  Dragan]{Shah19}
Rohin Shah, Dmitrii Krasheninnikov, Jordan Alexander, Pieter Abbeel, and Anca
  Dragan.
\newblock Preferences implicit in the state of the world.
\newblock In \emph{International Conference for Learning Representations
  (ICLR)}, 2019.

\bibitem[Taylor(2016)]{Taylor2016quant}
Jessica Taylor.
\newblock Quantilizers: A safer alternative to maximizers for limited
  optimization.
\newblock In \emph{AAAI Workshop on AI, Ethics, and Society}, pages 1--9, 2016.

\bibitem[Taylor et~al.(2016)Taylor, Yudkowsky, LaVictoire, and
  Critch]{Taylor16}
Jessica Taylor, Eliezer Yudkowsky, Patrick LaVictoire, and Andrew Critch.
\newblock Alignment for advanced machine learning systems.
\newblock Technical report, Machine Intelligence Research Institute, 2016.

\bibitem[Turchetta et~al.(2016)Turchetta, Berkenkamp, and Krause]{Turchetta16}
Matteo Turchetta, Felix Berkenkamp, and Andreas Krause.
\newblock Safe exploration in finite {M}arkov decision processes with
  {G}aussian processes.
\newblock In \emph{Neural Information Processing Systems (NIPS)}, pages
  4305--4313, 2016.

\bibitem[Turner et~al.(2019)Turner, Hadfield-Menell, and Tadepalli]{Turner19}
Alexander Turner, Dylan Hadfield-Menell, and Prasad Tadepalli.
\newblock Conservative agency via attainable utility preservation.
\newblock \emph{arXiv preprint arXiv:1902.09725}, 2019.

\bibitem[Warnell et~al.(2018)Warnell, Waytowich, Lawhern, and Stone]{Warnell18}
Garrett Warnell, Nicholas~R. Waytowich, Vernon Lawhern, and Peter Stone.
\newblock Deep {TAMER:} interactive agent shaping in high-dimensional state
  spaces.
\newblock In \emph{{AAAI} Conference on Artificial Intelligence}, 2018.

\bibitem[Zhang et~al.(2018)Zhang, Durfee, and Singh]{Zhang18}
Shun Zhang, Edmund~H. Durfee, and Satinder~P. Singh.
\newblock Minimax-regret querying on side effects for safe optimality in
  factored {Markov} decision processes.
\newblock In \emph{International Joint Conference on Artificial Intelligence
  {(IJCAI)}}, pages 4867--4873, 2018.

\bibitem[Ziebart et~al.(2008)Ziebart, Maas, Bagnell, and Dey]{Ziebart08}
Brian~D Ziebart, Andrew~L Maas, J~Andrew Bagnell, and Anind~K Dey.
\newblock Maximum entropy inverse reinforcement learning.
\newblock In \emph{AAAI}, pages 1433--1438, 2008.

\end{thebibliography}

\ifarxiv

\begin{appendices}

\newpage



\section{Relationship between discounted and undiscounted reachability}

\begin{proposition}\label{prop:limfin}
We define discounted reachability as follows:
\begin{equation}\label{def:fin}
\covfin{\tilde s}{s} := \max_\pi \expect [\gammar^{\ns{\pi}{\tilde s}{s}}].
\end{equation}
and undiscounted reachability as follows:
\begin{equation}\label{def:inf}
\covinf{\tilde s}{s} := \max_\pi P(\ns{\pi}{\tilde s}{s} < \infty)
\end{equation}

We show that for all $s, \tilde s$, as $\gammar \rightarrow 1$, discounted reachability approaches undiscounted reachability: 
$\lim_{\gammar \rightarrow 1} \covfin{\tilde s}{s} = \covinf{\tilde s}{s}$.
\end{proposition}

\begin{proof}
First we show for fixed $\pi$ that
\begin{align}
&\lim_{\gammar \rightarrow 1} \expect [\gammar^{\ns{\pi}{\tilde s}{s}}] \nonumber\\
=& \lim_{\gammar \rightarrow 1} P(\ns{\pi}{\tilde s}{s} < \infty) \expect[\gammar^{\ns{\pi}{\tilde s}{s}} | \ns{\pi}{\tilde s}{s} < \infty]
+ \lim_{\gammar \rightarrow 1}P(\ns{\pi}{\tilde s}{s} = \infty) \expect[\gammar^{\ns{\pi}{\tilde s}{s}} | \ns{\pi}{\tilde s}{s} = \infty] \nonumber\\
=& P(\ns{\pi}{\tilde s}{s} < \infty) \cdot 1 + P(\ns{\pi}{\tilde s}{s} = \infty) \cdot 0 \nonumber\\
=& P(\ns{\pi}{\tilde s}{s} < \infty). \label{eq:lemma1.1}
\end{align}

Now let $\pi(\gammar)$ be an optimal policy for that value of $\gammar$: $\pi(\gammar) := \arg\max_\pi  \expect [\gammar^{\ns{\pi}{\tilde s}{s}}]$. For any $\epsilon$, there is a $\tgammar$ such that both of the following hold:
\begin{align*}
\left| \expect [\tgammar^{\ns{\pi(\tgammar)}{\tilde s}{s}}] - P(\ns{\pi(\tgammar)}{\tilde s}{s} < \infty) \right| &<\epsilon \quad\text{(by equation \ref{eq:lemma1.1}) and}\\
\left| \expect [\tgammar^{\ns{\pi(\tgammar)}{\tilde s}{s}}] - \lim_{\gammar \rightarrow 1} \expect [\gammar^{\ns{\pi(\gammar)}{\tilde s}{s}}] \right| &< \epsilon \quad\text{(assuming the limit exists).}
\end{align*}

Thus, $| \lim_{\gammar \rightarrow 1} \expect [\gammar^{\ns{\pi(\gammar)}{\tilde s}{s}}] - P(\ns{\pi(\tgammar)}{\tilde s}{s} < \infty) | < 2\epsilon$. Taking $\epsilon \rightarrow 0$, we have 
\begin{equation}\label{eq:lemma1.2}
\lim_{\gammar \rightarrow 1} \covfin{\tilde s}{s} = \lim_{\gammar \rightarrow 1} \expect [\gammar^{\ns{\pi(\gammar)}{\tilde s}{s}}] = \lim_{\tgammar \rightarrow 1} P(\ns{\pi(\tgammar)}{\tilde s}{s} < \infty).
\end{equation}

Let $\tilde\pi = \arg\max_\pi  P(\ns{\pi}{\tilde s}{s} < \infty)$. Then,
\begin{align*}
\max_\pi P(\ns{\pi}{\tilde s}{s} < \infty) &=\lim_{\gammar \rightarrow 1} \expect [\gammar^{\ns{\tilde \pi}{\tilde s}{s}}] &&\text{(by equation \ref{eq:lemma1.1})} \\
&\leq  \lim_{\gammar \rightarrow 1} \expect [\gammar^{\ns{\pi(\gammar)}{\tilde s}{s}}] &&\text{(since $\pi(\gammar)$ is optimal for each $\gammar$)}
\end{align*}
Also,
\begin{align*}
\lim_{\gammar \rightarrow 1} \expect [\gammar^{\ns{\pi(\gammar)}{\tilde s}{s}}]
&= \lim_{\tilde \gammar \rightarrow 1} P(\ns{\pi(\tgammar)}{\tilde s}{s} < \infty) &&\text{(by equation \ref{eq:lemma1.2})}\\
&\leq \lim_{\tilde \gammar \rightarrow 1} P(\ns{\tilde\pi}{\tilde s}{s} < \infty) \\
&= \max_\pi P(\ns{\pi}{\tilde s}{s} < \infty)
\end{align*}

Thus they are equal, which completes the proof.
\end{proof}

\section{Reachability variant for large state spaces using a measure of similarity between states}

In large state spaces, the agent might not be able to reach the given state $s$, but able to reach states that are similar to $s$ according to some distance measure $\delta$. We will now extend our previous definitions to this case by defining 
\emph{similarity-based reachability}:
\begin{align}
\text{Discounted: } \covsimfin{\tilde s}{s} & := \max_\pi \sum_{t=0}^\infty (1-\gammar)\gammar^t \expect [e^{-\diff{\simstate{\pi}}{s}}] \label{def:sim}\\
\text{Undiscounted: } \covsiminf{\tilde s}{s} & := \max_\pi \lim_{t\rightarrow \infty} \expect[e^{-\diff{\simstate{\pi}}{s}}] \label{def:siminf}
\end{align}
where $\simstate{\pi}$ is the state that the agent is in after following policy $\pi$ for $t$ steps starting from $\tilde s$. Discounted similarity-based reachability is proportional to the value function of the optimal policy $\pi$ for an agent that gets reward $e^{-\diff{\tilde s}{s}}$ in state $\tilde s$ (which rewards the agent for going to states $\tilde s$ that are similar to $s$) and uses a discount factor of $\gammar$. Undiscounted similarity-based reachability represents the highest reward the agent could attain in the limit by going to states as similar to $s$ as possible.

\begin{proposition}\label{prop:limsim}
For all $s, \tilde s, \delta$, as $\gammar \rightarrow 1$, similarity-based discounted reachability \eqref{def:sim} approaches similarity-based undiscounted reachability \eqref{def:siminf}:
$\lim_{\gammar \rightarrow 1} \covsimfin{\tilde s}{s} = \covsiminf{\tilde s}{s}.$
\end{proposition}

\begin{proof}
First we show for fixed $\pi$ that if $\lim_{t\rightarrow \infty} \expect[e^{-\diff{\simstate{\pi}}{s}}]$ exists, then 
\begin{equation}\label{eq:lemma2.1}
\lim_{\gammar \rightarrow 1} \sum_{t=0}^\infty (1-\gammar)\gammar^t \expect [e^{-\diff{\simstate{\pi}}{s}}] = \lim_{t\rightarrow \infty} \expect[e^{-\diff{\simstate{\pi}}{s}}]
\end{equation}

Let $x_t = \expect [e^{-\diff{\simstate{\pi}}{s}}]- \lim_{t\rightarrow \infty} \expect[e^{-\diff{\simstate{\pi}}{s}}]$. Since $x_t \rightarrow 0$ as $t \rightarrow \infty$, for any $\epsilon$ we can find a large enough $t_\epsilon$ such that $|x_t| \leq \epsilon$, $\forall t>t_\epsilon$. Then, we have
\begin{align*} 
\lim_{\gammar \rightarrow 1} \sum_{t=0}^\infty (1-\gammar)\gammar^t x_t 
&= \lim_{\gammar \rightarrow 1} \sum_{t=0}^{t_\epsilon-1} (1-\gammar)\gammar^t x_t + \lim_{\gammar \rightarrow 1} \sum_{t=t_\epsilon}^\infty  (1-\gammar)\gammar^t x_t \\
&\leq \lim_{\gammar \rightarrow 1} (1-\gammar)\cdot \lim_{\gammar \rightarrow 1} \sum_{t=0}^{t_\epsilon-1} \gammar^t x_t + \lim_{\gammar \rightarrow 1}\sum_{t=t_\epsilon}^\infty  (1-\gammar)\gammar^t \epsilon\\
&= 0 + \epsilon \lim_{\gammar \rightarrow 1} \gammar^{t_\epsilon} \\
&= \epsilon.
\end{align*}

Similarly, we can show that $\lim_{\gammar \rightarrow 1} \sum_{t=0}^\infty (1-\gammar)\gammar^t x_t \geq -\epsilon$. Since this holds for all $\epsilon$, $$\lim_{\gammar \rightarrow 1} \sum_{t=0}^\infty (1-\gammar)\gammar^t x_t = 0$$
which is equivalent to equation \ref{eq:lemma2.1}.

Now let $\pi(\gammar)$ be an optimal policy for that value of $\gammar$: $\pi(\gammar) := \arg\max_\pi  \sum_{t=0}^\infty (1-\gammar)\gammar^t \expect [e^{-\diff{\simstate{\pi}}{s}}]$.
For any $\epsilon$, there is a $\tgammar$ such that both of the following hold:
\begin{align*}
\left|\sum_{t=0}^\infty (1-\tgammar){\tgammar}^t \expect [e^{-\diff{\simstate{\pi(\tgammar)}}{s}}] - \lim_{t\rightarrow \infty} \expect[e^{-\diff{\simstate{\pi(\tgammar)}}{s}}] \right| &<\epsilon \quad\text{(by equation \ref{eq:lemma2.1}) and}\\
\left|\sum_{t=0}^\infty (1-\tgammar){\tgammar}^t \expect [e^{-\diff{\simstate{\pi(\tgammar)}}{s}}] - \lim_{\gammar \rightarrow 1} \sum_{t=0}^\infty (1-\gammar)\gammar^t \expect [e^{-\diff{\simstate{\pi(\gammar)}}{s}}]\right| &< \epsilon \quad\text{(assuming the limit exists).}
\end{align*}

Thus, $|\lim_{\gammar \rightarrow 1} \sum_{t=0}^\infty (1-\gammar)\gammar^t \expect [e^{-\diff{\simstate{\pi(\gammar)}}{s}}] - \lim_{t\rightarrow \infty} \expect[e^{-\diff{\simstate{\pi(\tgammar)}}{s}}]| < 2\epsilon$. Taking $\epsilon \rightarrow 0$, we have 
\begin{equation}\label{eq:lemma2.2}
\lim_{\gammar \rightarrow 1} \covsimfin{\tilde s}{s} = \lim_{\gammar \rightarrow 1} \sum_{t=0}^\infty (1-\gammar)\gammar^t \expect [e^{-\diff{\simstate{\pi(\gammar)}}{s}}] = \lim_{\tgammar \rightarrow 1}\lim_{t\rightarrow \infty} \expect[e^{-\diff{\simstate{\pi(\tgammar)}}{s}}].
\end{equation}

Let $\tilde\pi = \arg\max_\pi \lim_{t\rightarrow \infty} \expect[e^{-\diff{\simstate{\pi}}{s}}]$ be the optimal policy for the similarity-based undiscounted reachability. Then,
\begin{align*}
\max_\pi \lim_{t\rightarrow \infty} \expect[e^{-\diff{\simstate{\pi}}{s}}] &=\lim_{\gammar \rightarrow 1} \sum_{t=0}^\infty (1-\gammar)\gammar^t \expect [e^{-\diff{\simstate{\tilde\pi}}{s}}] &&\text{(by equation \ref{eq:lemma2.1})} \\
&\leq  \lim_{\gammar \rightarrow 1} \sum_{t=0}^\infty (1-\gammar)\gammar^t \expect [e^{-\diff{\simstate{\pi(\gammar)}}{s}}] &&\text{(since $\pi(\gammar)$ is optimal for each $\gammar$)} \\
&= \lim_{\gammar \rightarrow 1}\lim_{t\rightarrow \infty} \expect[e^{-\diff{\simstate{\pi(\gammar)}}{s}}] &&\text{(by equation \ref{eq:lemma2.2})}\\
&\leq \max_\pi \lim_{t\rightarrow \infty} \expect[e^{-\diff{\simstate{\pi}}{s}}] 
\end{align*}

Thus, equality holds throughout, which completes the proof.
\end{proof}

\begin{proposition}\label{prop:indicator}
Let the indicator distance $\delta_\mathbb{I}$ be a distance measure with $\diffind{s_i}{s_j} = 0$ if $s_i=s_j$ and $\infty$ otherwise (so it only matters whether the exact target state is reachable). Then for all $s, \tilde s, \gammar$, 
\begin{itemize}
\item similarity-based discounted reachability \eqref{def:sim} is equivalent to discounted reachability \eqref{def:fin}:
$\covsimfinind{\tilde s}{s} = \covfin{\tilde s}{s}$,
\item similarity-based undiscounted reachability \eqref{def:siminf} is equivalent to undiscounted reachability \eqref{def:inf}:\\
$\covsiminfind{\tilde s}{s} = \covinf{\tilde s}{s}$.
\end{itemize}
\end{proposition}
\begin{proof}
$\begin{aligned}[t]
\covsimfinind{\tilde s}{s} &= \max_\pi \sum_{t=0}^\infty (1-\gammar)\gammar^t \expect [e^{-\diff{\simstate{\pi}}{s}}] \\
&= \max_\pi \left( \expect \left[ \sum_{t=0}^{\ns{\pi}{\tilde s}{s}-1} (1-\gammar)\gammar^t e^{-\infty}\right] + \expect \left[ \sum_{t=\ns{\pi}{\tilde s}{s}}^\infty (1-\gammar)\gammar^t e^0 \right]\right) \\
&= \max_\pi \left( 0 + \expect \left[\gammar^{\ns{\pi}{\tilde s}{s}}(1-\gammar) \sum_{t=0}^\infty \gammar^t\right]\right)  \\
&= \max_\pi \expect \left[\gammar^{\ns{\pi}{\tilde s}{s}} \cdot 1 \right]\\
&= \covfin{\tilde s}{s}.\\
\covsiminfind{\tilde s}{s}
&= \max_\pi \lim_{t\rightarrow \infty} \expect[e^{-\diff{\simstate{\pi}}{s}}] \\
&= \max_\pi (P(\ns{\pi}{\tilde s}{s} < \infty) e^0 + P (\ns{\pi}{\tilde s}{s} = \infty) e^{-\infty}) \\
&= \max_\pi P(\ns{\pi}{\tilde s}{s} < \infty)\\
&= \covinf{\tilde s}{s}. && \mbox{\qedhere}
\end{aligned}$
\end{proof}

We can represent the relationships between the reachability definitions as follows:
\begin{alignat*}{2}
R_{\gammar,\delta}\; \eqref{def:sim} &\xrightarrow{\gammar \rightarrow 1 (\text{Prop } \ref{prop:limsim})} & R_{1,\delta} \; \eqref{def:siminf} \quad\quad\quad\\
{\scriptstyle\delta = \delta_\mathbb{I} (\text{Prop } \ref{prop:indicator})} \downarrow \quad & & \downarrow {\scriptstyle\delta = \delta_\mathbb{I} (\text{Prop } \ref{prop:indicator})} \\
R_{\gammar} \; \eqref{def:fin} &\xrightarrow{\gammar \rightarrow 1 (\text{Prop } \ref{prop:limfin})} & R_1\; \eqref{def:inf}\quad\quad\quad
\end{alignat*}


\section{Relative reachability computation example}\label{app:vases}

\begin{example}\label{ex:vases}
A variation on Example \ref{ex:vase}, where the environment contains two vases (vase 1 and vase 2) and the agent's goal is to do nothing. The agent can take action $b_i$ to break vase $i$. The MDP is shown in Figure \ref{fig:vases}.
\end{example}

\begin{figure}[ht]
  \begin{center}
    \begin{tikzpicture}[auto]
      \node[draw,circle] (s1) at (0, .9) {$s_1$};
      \node[draw,circle] (s2) at (-.9,0) {$s_2$};
      \node[draw,circle] (s4) at (0,-.9) {$s_4$};
      \node[draw,circle] (s3) at (.9, 0) {$s_3$};
      \draw (s2) edge[<-,>=latex] node{$b_1$} (s1);
      \draw (s1) edge[->,>=latex] node{$b_2$} (s3);
      \draw (s4) edge[<-,>=latex] node{$b_2$} (s2);
      \draw (s3) edge[->,>=latex] node{$b_1$} (s4);
      \draw (s1) edge[->, >=latex, loop below] (s1);
      \draw (s2) edge[->, >=latex, loop right] (s2);
      \draw (s3) edge[->, >=latex, loop left]  (s3);
      \draw (s4) edge[->, >=latex, loop above] (s4);
      \node[above = .5mm of s1] {no vases broken};
      \node[left  = 1mm of s2] {vase 1 broken};
      \node[below = .5mm of s4] {vases 1 and 2 broken};
      \node[right = 1mm of s3] {vase 2 broken};
    \end{tikzpicture}
  \end{center}
  \caption{Transitions between states when breaking vases in Example \ref{ex:vases}.}
  \label{fig:vases}
\end{figure}
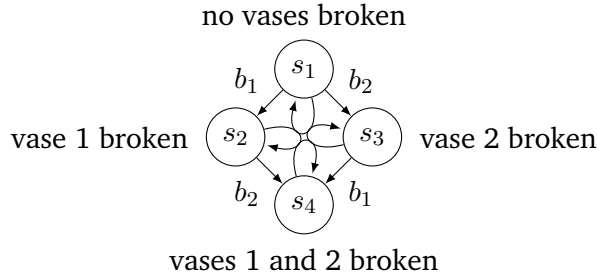

We compute the relative reachability of different states from $s_2$ using undiscounted reachability:
\begin{align*}
\devc{s_2}{s_3} =& \frac{1}{4} \sum_{k=1}^4 \tr{\covinf{s_3}{s_k} - \covinf{s_2}{s_k}} \\
&= \frac{1}{4} ( \cancel{\tr{0-0}} + \cancel{\tr{0-1}} + \tr{1-0} + \cancel{\tr{1-1}}) \\
&= \frac{1}{4} , \\ 
\devc{s_2}{s_1} =& \frac{1}{4}  \sum_{k=1}^4 \tr{\covinf{s_1}{s_k} - \covinf{s_2}{s_k}} \\
&= \frac{1}{4} ( \tr{1-0} + \cancel{\tr{1-1}} + \tr{1-0} + \cancel{\tr{1-1}}) \\
&= \frac{1}{2},
\end{align*}
where $\covinf{s_i}{s_k}$ is 1 if $s_k$ is reachable from $s_i$ and 0 otherwise.

Now we compute the relative reachability of different states from $s_2$ using discounted reachability:
\begin{align*}
\devc{s_2}{s_3} =& \frac{1}{4}  \sum_{k=1}^4 \tr{\covfin{s_3}{s_k} - \covfin{s_2}{s_k}} \\
=& \frac{1}{4} ( \tr{\gammar^\infty - \gammar^\infty} + \tr{\gammar^\infty - \gammar^0} + \tr{\gammar^0 - \gammar^\infty} + \tr{\gammar^1 - \gammar^1})\\
=& \frac{1}{4} (\cancel{\tr{0-0}} + \cancel{\tr{0-1}} + \tr{1-0} + \cancel{\tr{\gammar-\gammar}} )\\
=& \frac{1}{4} ,\\ 
\devc{s_2}{s_1} =& \frac{1}{4} \sum_{k=1}^4 \tr{\covfin{s_1}{s_k} - \covfin{s_2}{s_k}} \\
=& \frac{1}{4} (\tr{\gammar^0 - \gammar^\infty} + \tr{\gammar^1 - \gammar^0} + \tr{\gammar^1 - \gammar^\infty} + \tr{\gammar^2 - \gammar^1})\\
=& \frac{1}{4} (\tr{1-0} + \cancel{\tr{\gammar-1}} + \tr{\gammar-0} + \cancel{\tr{\gammar^2-\gammar}})\\
=& \frac{1}{4} (1+\gammar) \xrightarrow{\gammar\rightarrow1} \frac{1}{2} .
\end{align*}

\end{appendices}

\ifextensions

\pagenumbering{gobble}

\vfill\pagebreak

\subfile{extensions/ablation.tex}

\vfill\pagebreak

\subfile{extensions/approximations.tex}

\fi

\fi

\end{document}